\documentclass[11pt]{article}
\usepackage{algorithm}      
\usepackage{algpseudocode}  
\usepackage{amsmath}        
\usepackage[margin=1in]{geometry}
\usepackage{graphicx}
\usepackage{booktabs}
\usepackage{multirow}
\usepackage{amsmath,amssymb}
\usepackage{microtype}
\usepackage{siunitx}
\usepackage{array}
\usepackage{xurl}
\usepackage[hidelinks]{hyperref}
\usepackage[backend=biber,style=numeric-comp,sorting=none,maxbibnames=30]{biblatex}
\AtBeginDocument{\let\cite\supercite}
\usepackage{authblk}
\usepackage{lineno}
\usepackage{caption}
\usepackage{booktabs}
\usepackage{subcaption}
\usepackage{tikz}
\usetikzlibrary{arrows.meta,positioning,fit}
\title{Learning Pathology-Sensitive Representations for Unsupervised Retinal OCT Anomaly Detection}
\author[1]{Tania Haghighi}
\author[1]{Sina Gholami}
\author[1]{Behafarin Emam}
\author[2]{Jennifer I Lim}
\author[3]{Theodore Leng}
\author[4]{Atalie C Thompson}
\author[1]{Hamed Tabkhi}
\author[1]{Minhaj Nur Alam\thanks{Corresponding author: \texttt{malam8@charlotte.edu}}}
\affil[1]{University of North Carolina at Charlotte, Charlotte, North Carolina, USA}
\affil[2]{University of Illinois Chicago, Chicago, IL, USA}
\affil[3]{Byers Eye Institute at Stanford, Stanford University School of Medicine, Stanford, CA, USA}
\affil[4]{Surgical Ophthalmology, Atrium Health Wake Forest Baptist, Winston–Salem, NC, USA}
\date{}

\begin{document}
\maketitle
\begin{abstract}
Deep learning approaches for retinal optical coherence tomography (OCT) predominantly rely on predefined disease categories, limiting their ability to detect rare or previously unseen abnormalities. While visual foundation models provide transferable representations, their effectiveness in identifying pathological deviations from normal retinal anatomy remains insufficiently understood. In this work, we introduce an unsupervised framework for retinal OCT anomaly detection and localization that adapts pretrained visual representations to normal retinal anatomy through anatomy-aware reconstruction and contrastive triplet learning guided by synthetic structural perturbations, without requiring disease-specific annotations. We systematically investigate the influence of pretraining strategies, anatomy-aware representation adaptation, and anomaly scoring mechanisms on the detection and localization of unseen retinal abnormalities. Our findings demonstrate that adapting pretrained representations to normal retinal structure improves anomaly sensitivity, while pixel-space reconstruction errors and feature-space deviations provide complementary signals for identifying pathological changes. These findings underscore the importance of anatomy-aware representation learning and offer insights into developing generalizable retinal anomaly detection frameworks beyond predefined disease categories.
\end{abstract}

\section{Introduction}

Optical coherence tomography (OCT) has become central to the diagnosis and monitoring of retinal diseases, with approximately 30 million scans performed worldwide each year \cite{fujimoto2016oct}. Deep learning has achieved expert-level performance in OCT analysis when clinical tasks are defined by known diagnostic categories or annotated pathological findings \cite{schmidterfurth2018ai,ting2019ai}, including disease classification \cite{kermany2018identifying}, referral prediction \cite{defauw2018clinically}, and fluid quantification \cite{schlegl2018fluid}. Several ophthalmic artificial intelligence systems have also been approved as medical devices \cite{ong2025devices}. More recently, visual foundation models pretrained on large image collections have demonstrated transferable representations for ophthalmic classification, prognosis, and segmentation with reduced reliance on task-specific annotations \cite{zhou2023retfound,qiu2024visionfm,morano2025mirage,gholami2025distributed,zhang2026focus}, reflecting a broader movement toward generalist medical artificial intelligence \cite{moor2023generalist,zhang2024challenges}. Despite these advances, most existing approaches remain dependent on predefined disease categories or annotated abnormalities, limiting their ability to recognize pathological findings that are rare, heterogeneous, or absent from the training data.

This limitation is particularly important in retinal OCT, where pathological changes can manifest as subtle disruptions of anatomical structures, coexist with other abnormalities, or differ substantially from previously observed disease patterns. Supervised models can achieve strong performance within their predefined label spaces but remain vulnerable to underrepresented conditions \cite{burlina2020lowshot} and previously unseen abnormalities \cite{roy2022longtail}. Consequently, reliable retinal analysis requires moving beyond disease-specific recognition toward learning representations that can distinguish normal anatomical variation from potentially pathological deviations. This raises a fundamental question: can a model learn pathology-sensitive representations capable of detecting and localizing previously unseen retinal abnormalities without requiring annotations for individual diseases? Addressing this question could support more generalizable retinal analysis systems that remain effective across heterogeneous clinical populations and imaging conditions.

Visual foundation models offer a promising starting point for learning such representations. Through large-scale pretraining, these models acquire transferable visual features that can support downstream tasks with limited supervision \cite{zhou2023retfound,oquab2024dinov2,simeoni2025dinov3}. Pretrained representations have also demonstrated value for anomaly detection in natural, industrial, and medical images \cite{reiss2021panda,roth2022patchcore,huang2024mvfa}. However, the representations that support semantic recognition are not necessarily optimized for detecting subtle pathological deviations. Pretraining objectives often encourage invariance to local appearance variations, whereas retinal anomaly detection requires sensitivity to small structural changes that may indicate disease. Moreover, the relative suitability of different pretraining strategies remains unresolved. Studies of supervised ophthalmic tasks have reported advantages for retina-specific foundation models \cite{zhou2023retfound,zhou2025generalist}, while others have found that general-purpose or natural-image pretrained models can match or outperform domain-specific alternatives \cite{hou2026natural,zoellin2026benchmark}, consistent with broader evidence of effective transfer from natural-image representations to medical imaging \cite{huix2024natural}. Importantly, strong performance on supervised ophthalmic benchmarks does not necessarily imply sensitivity to unseen pathological abnormalities. These observations motivate a more fundamental investigation of whether pretrained visual representations contain sufficient anatomical information for anomaly detection and, critically, how they can be adapted to become sensitive to pathological deviations from normal retinal structure.

Unsupervised anomaly detection provides a natural framework for investigating this problem. Rather than learning predefined disease categories, unsupervised methods characterize normal anatomical patterns and identify deviations that are inconsistent with those patterns, reducing dependence on exhaustive disease annotations \cite{fernando2021survey,cai2025medianomaly}. Reconstruction-based approaches, including autoencoders and generative adversarial networks, identify abnormalities through discrepancies between observed images and reconstructions learned from normal data \cite{schlegl2017unsupervised,schlegl2019fano,baur2021autoencoders}. Feature-based methods estimate anomalies by measuring deviations from normal feature distributions \cite{defard2021padim,roth2022patchcore}, while self-supervised approaches use synthetic anomalies to encourage sensitivity to abnormal patterns without requiring annotated pathological examples \cite{li2021cutpaste,zavrtanik2021draem,schluter2022nsa,tian2023pmsacl}. In retinal OCT, these approaches have demonstrated the ability to identify potential disease markers \cite{seebock2019disease}, detect diverse pathological changes \cite{seebock2020uncertainty,zhou2020pnet,aresta2025kd}, and localize abnormalities without lesion-level supervision \cite{schlegl2017unsupervised,seebock2019disease}. Nevertheless, systematic evaluations have revealed substantial variability in performance across methods and datasets \cite{baur2021autoencoders,lagogiannis2024deep,cai2025medianomaly,bao2024bmad}. These findings suggest that the effectiveness of unsupervised anomaly detection depends not only on the anomaly scoring mechanism but also on the underlying representation and its ability to preserve anatomically meaningful deviations.

A central challenge, therefore, is learning representations that capture normal retinal anatomy while remaining sensitive to structural changes that may indicate pathology. Retinal OCT exhibits highly organized anatomical patterns, including layered tissue structures whose spatial relationships provide important cues for identifying abnormalities. Representations that preserve these relationships may offer a stronger basis for distinguishing pathological deviations from normal variation. Anatomy-aware reconstruction can encourage the model to encode retinal structure, while contrastive learning with synthetic structural perturbations can explicitly promote sensitivity to anatomically meaningful changes without requiring real pathological annotations. Together, these mechanisms provide a means of adapting pretrained visual features toward pathology-sensitive representations. Furthermore, pixel-space reconstruction discrepancies and feature-space deviations may capture complementary manifestations of abnormalities, suggesting that representation adaptation and anomaly estimation should be investigated jointly rather than treated as independent design choices.

Motivated by these considerations, we investigate how pretrained visual representations can be adapted to learn pathology-sensitive features for unsupervised retinal OCT anomaly detection and localization. Our framework integrates pretrained visual encoders with anatomy-aware reconstruction and contrastive triplet learning guided by synthetic structural perturbations, without requiring disease-specific annotations. We systematically examine the influence of pretraining strategies, anatomical adaptation, and anomaly scoring mechanisms on the detection of previously unseen abnormalities. In particular, we investigate whether representations optimized for general visual recognition can be transformed into effective anomaly-sensitive representations through adaptation to normal retinal anatomy, and how pixel-space and feature-space anomaly signals contribute to detection and localization performance.

The main contributions of this study are:

\begin{itemize}

\item We introduce a framework for learning pathology-sensitive representations for unsupervised retinal OCT anomaly detection and localization. The framework integrates pretrained visual representations with anatomy-aware reconstruction and contrastive triplet learning guided by synthetic structural perturbations, enabling the identification of pathological deviations without disease-specific annotations.

\item We systematically investigate how pretraining strategies and anatomy-aware representation adaptation influence sensitivity to previously unseen retinal abnormalities. Through controlled experiments, we examine the roles of reconstruction objectives, retinal region weighting, synthetic perturbation diversity, and normal training data availability in learning representations that distinguish anatomical variations from pathological deviations.

\item We investigate the complementary roles of pixel-space reconstruction errors and feature-space deviations in detecting and localizing retinal abnormalities. We evaluate how anomaly scoring mechanisms influence detection performance, localization accuracy, and cross-device generalization without additional training or pathological annotations.

\end{itemize}

\section{Results}


\subsection{Study overview}
\begin{figure}[t]
    \centering
    \includegraphics[width=\linewidth]{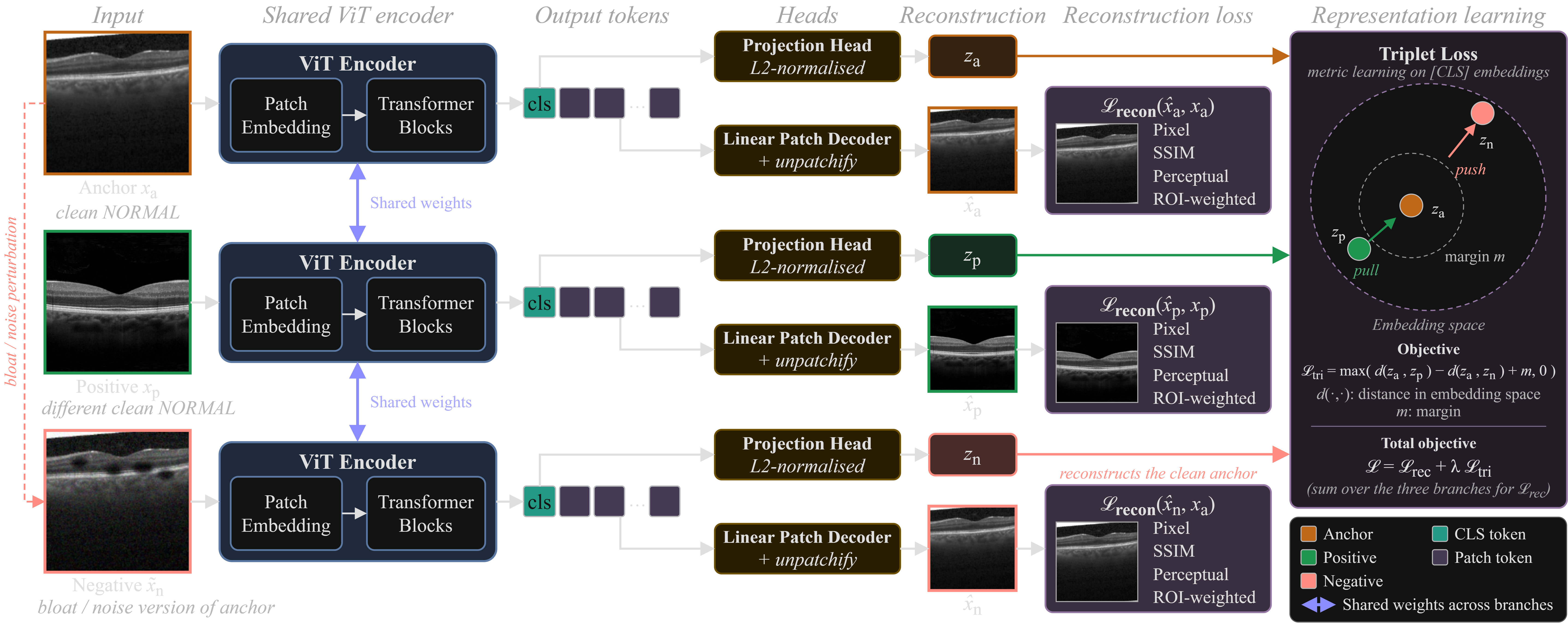}
    \caption{
    Overview of the proposed representation-learning framework. 
    A shared ViT encoder processes a clean anchor OCT B-scan, a different clean normal positive, and a perturbed version of the anchor as the negative. 
    CLS tokens are mapped to L2-normalized embeddings and optimized using triplet loss, while patch tokens are reconstructed through a linear patch decoder using the reconstruction objective.
    }
    \label{fig:framework}
\end{figure}

We developed an anatomy-aware, triplet-guided framework that adapts pretrained vision transformers for unsupervised retinal OCT anomaly detection and localization (Fig.~\ref{fig:framework}). The framework learns to recognize deviations from normal retinal anatomy without requiring pathological training images or disease-specific annotations. During training, three images are used: an anchor from a normal retina, a positive from a different normal retina, and a negative generated by introducing  perturbations into the anchor. All three images pass through a shared encoder, which produces representations used for two complementary learning objectives. A linear patch decoder reconstructs the normal images and learns to recover the clean anchor from its perturbed version, with reconstruction losses emphasizing retinal structures. Meanwhile, a triplet objective encourages similar representations for normal images while separating synthetically altered images from them. Together, these objectives teach the model to reconstruct normal retinal anatomy while remaining sensitive to abnormal structural changes. After training, abnormalities are identified through reconstruction differences or deviations from normal feature representations. We evaluated the models using normal images from Kermany \cite{kermany2018identifying} for training, its held-out test set for image-level detection, and RETOUCH \cite{bogunovic2019retouch} for pixel-level localization without additional training. We further examined the effects of training objectives, data availability, anomaly scoring, and synthetic perturbations. Additional methodological details are provided in the Methods section.

\subsection{Datasets}
We used two public OCT datasets that play different roles in the study. The Kermany dataset was used to train and validate all models and to evaluate image-level detection. RETOUCH was used only for testing, to evaluate whether the models can localize pathology at the pixel level on images from devices they had not seen during training.
 
The Kermany dataset\cite{kermany2018identifying} contains macular B-scans acquired with Spectralis OCT (Heidelberg Engineering), each labeled as normal, CNV, DME or drusen. It is well suited to our setting for two reasons. It contains a large number of normal B-scans, which allows models to be trained on normal images alone, and its three disease classes differ in appearance, which allows detection to be tested across different types of structural change. We used the same data partition as in our previous study\cite{haghighi2026anatomy}. The training set contained 48,286 normal B-scans. The validation set contained 1,007 normal B-scans and was used to select checkpoints and to set decision thresholds using normal and synthetically perturbed validation images, so that no pathological image was involved in either step. The test set contained 2,019 B-scans, approximately half normal and half abnormal, with CNV, DME and drusen equally represented among the abnormal images. Disease labels were used only to evaluate the models.
 
Because the Kermany dataset provides labels at the image level only, it cannot show whether a model detects an abnormal image for the right reason, that is, whether the anomaly signal comes from the lesion itself. For this purpose we used RETOUCH\cite{bogunovic2019retouch}, which contains OCT volumes from patients with neovascular age-related macular degeneration or retinal vein occlusion, with expert pixel-level annotations of IRF, SRF and PED. RETOUCH images were acquired with three devices (Cirrus, Spectralis and Topcon), so the dataset also tests how well the models generalize beyond the device used for training. We used the RETOUCH validation set, for which reference annotations are available. No RETOUCH image or annotation was used for training, model selection or threshold calibration.

\subsection{Anomaly detection}
Image-level anomaly detection was evaluated on the Kermany test set, which contains normal B-scans and B-scans with choroidal neovascularization (CNV), diabetic macular edema (DME) or drusen. The complete models of the three encoders were compared with a ConvNeXt-based reconstruction model, a variational autoencoder (VAE), VQ-VAE, VQGAN, f-AnoGAN and an anatomy-aware VQGAN \cite{haghighi2026anatomy}. We report area under the receiver operating characteristic curve (AUROC), together with F1 score and accuracy at the operating threshold (Table~\ref{tab:detection}).

\begin{table}[t]
\centering
\caption{Image-level anomaly detection on the Kermany test set. F1 score and accuracy are computed at the operating threshold. Best values are shown in bold.}
\label{tab:detection}
\begin{tabular}{lccc}
\toprule
Method & AUROC & F1 (\%) & Accuracy (\%) \\
\midrule
ConvNeXt            & 0.716 & 69.7 & 62.8 \\
VAE                 & 0.685 & 69.6 & 60.0 \\
VQ-VAE              & 0.737 & 71.4 & 65.5 \\
VQGAN               & 0.747 & 71.3 & 67.6 \\
f-AnoGAN            & 0.763 & 71.1 & 68.4 \\
\midrule
(Ours) VQGAN & 0.799 & 75.2 & 71.7 \\
(Ours) DINOv3              & 0.719 & 63.2 & 70.5 \\
(Ours) RETFound            & 0.859 & 82.7 & 84.2 \\
(Ours) ViT        & \textbf{0.921} & \textbf{85.9} & \textbf{87.3} \\
\bottomrule
\end{tabular}
\end{table}

\subsection{Anomaly localization}
Pixel-level localization was evaluated on the RETOUCH validation set, with the expert annotations of intraretinal fluid (IRF), subretinal fluid (SRF) and pigment epithelial detachment (PED) as the reference. The models from the detection experiment were applied to RETOUCH without further training, and their anomaly maps were compared with the annotations using Dice and mean intersection over union (mIoU)(Table~\ref{tab:localization}).

\begin{table}[t]
\centering
\caption{Unsupervised anomaly localization on RETOUCH. All models were trained on Kermany only and applied to RETOUCH without further training. mIoU is foreground-only (background excluded from the class average). Best values are shown in bold.}
\label{tab:localization}
\begin{tabular}{lcc}
\toprule
Method & Dice & mIoU \\
\midrule
ConvNeXt            & 0.181 & 0.100 \\
VAE                 & 0.123 & 0.066 \\
VQ-VAE              & 0.185 & 0.102 \\
VQGAN               & 0.145 & 0.078 \\
\midrule
(Ours) VQGAN & 0.200 & 0.111 \\
(Ours) DINOv3              & 0.140 & 0.075 \\
(Ours) RETFound            & 0.220 & 0.124 \\
(Ours) ViT        & \textbf{0.242} & \textbf{0.138} \\
\bottomrule
\end{tabular}
\end{table}

\subsection{Ablation studies}
We performed four ablation studies, examining the training objective, the amount of normal training data, the anomaly scoring strategy and the perturbation strategy. Each was evaluated for detection on Kermany and for localization on RETOUCH.

\subsubsection{Training objective}
To assess the contribution of each part of the training objective, we trained each encoder in four configurations of increasing adaptation. In the first, the encoder was frozen and only the linear reconstruction head was trained with the reconstruction loss. In the second, the encoder and reconstruction head were jointly trained using only the reconstruction loss. In the third, the encoder and head were trained jointly with the reconstruction loss and the triplet objective. In the fourth, retinal region-of-interest (ROI) weighting was added to the pixel loss. Results are shown in Table~\ref{tab:objective}.

\begin{table*}[t]
\centering
\small
\caption{Effect of the training objective. Frozen: the pretrained encoder is fixed and only the linear reconstruction head is trained. + Reconstruction: the encoder and head are trained jointly using reconstruction loss only. + Triplet: the encoder and head are trained jointly with the reconstruction and triplet objectives. + ROI: retinal region-of-interest (ROI) weighting is added to the pixel loss (complete model). Image-level metrics are computed on Kermany and localization metrics on RETOUCH. mIoU is foreground-only (background excluded from the class average).}
\label{tab:objective}
\begin{tabular}{llccccccc}
\toprule
Encoder & Configuration & AUROC & F1 (\%) & Acc. (\%) & Prec. (\%) & Rec. (\%) & Dice & mIoU \\
\midrule
DINOv3       & Frozen           & 0.440 & 13.0 & 52.9 & 86.0 & 7.0 & 0.012 & 0.006 \\
             & + Reconstruction & 0.510 & 21.0 & 54.8 & 83.0 & 12.0 & 0.023 & 0.012 \\
             & + Triplet        & 0.834 & 71.6 & 76.9 & 91.7 & 58.7 & 0.074 & 0.038 \\
             & + ROI            & 0.719 & 63.2 & 70.5 & 83.0 & 51.0 & 0.140 & 0.075 \\
\midrule
RETFound     & Frozen           & 0.598 & 13.4 & 52.6 & 71.2 & 7.4 & 0.030 & 0.015 \\
             & + Reconstruction & 0.640 & 26.4 & 55.3 & 75.0 & 16.0 & 0.051 & 0.026 \\
             & + Triplet        & 0.875 & 78.2 & 81.3 & 92.9 & 67.5 & 0.142 & 0.076 \\
             & + ROI            & 0.859 & 82.7 & 84.2 & 90.4 & 76.3 & 0.220 & 0.124 \\
\midrule
ImageNet ViT & Frozen           & 0.577 & 15.7 & 52.6 & 66.4 & 8.9 & 0.020 & 0.010 \\
             & + Reconstruction & 0.625 & 29.1 & 56.2 & 76.0 & 18.0 & 0.039 & 0.020 \\
             & + Triplet        & 0.910 & 81.5 & 84.0 & \textbf{95.5} & 71.2 & 0.091 & 0.048 \\
             & + ROI            & \textbf{0.921} & \textbf{85.9} & \textbf{87.3} & 89.1 & \textbf{84.7} & \textbf{0.242} & \textbf{0.138} \\
\bottomrule
\end{tabular}
\end{table*}

\subsubsection{Amount of training data}
\begin{figure}[t]
\centering
\includegraphics[width=\linewidth]{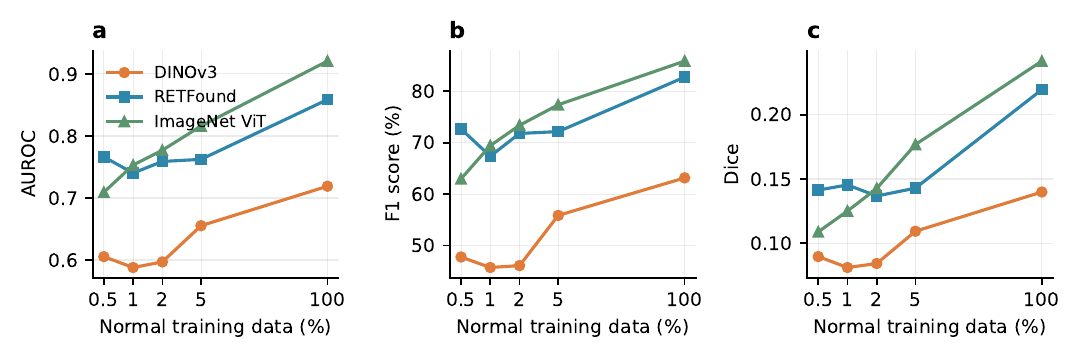}
\caption{Effect of the amount of normal training data on the complete models. \textbf{a}, AUROC and \textbf{b}, F1 score for image-level detection on Kermany. \textbf{c}, Dice for anomaly localization on RETOUCH. The x axis shows the percentage of the normal training set used for training, on a logarithmic scale.}
\label{fig:data_sensitivity}
\end{figure}

To assess how much normal data is required, we trained the complete models on random subsets containing 0.5\%, 1\%, 2\%, and 5\% of the normal training set and compared them with models trained on the full training set. All other settings were kept fixed. As shown in Table~\ref{tab:data} and Fig.~\ref{fig:data_sensitivity}, performance generally improved with increasing training data, although the three encoders showed different patterns. RETFound performed best with only 0.5\% of the training data, achieving an AUROC of 0.766 and a Dice score of 0.141, suggesting an advantage of retinal-domain pretraining when normal training samples are limited. In contrast, the ImageNet-pretrained ViT showed more consistent improvements as additional data became available, ultimately achieving the highest AUROC (0.921) and Dice score (0.242) with the full training set. DINOv3 remained the weakest performer across all training fractions. These findings indicate that the amount of normal training data influences the relative effectiveness of different pretrained representations, with RETFound offering an advantage under highly limited data availability and ImageNet-pretrained ViT benefiting more from larger training sets.

\begin{table*}[t]
\centering
\small
\caption{Effect of the amount of normal training data on the complete models. Percentages refer to the fraction of the normal training set used for training. Image-level metrics are computed on Kermany and localization metrics on RETOUCH. mIoU is foreground-only (background excluded from the class average). Bold values indicate the best performance for each metric at a given training-data percentage.}
\label{tab:data}
\begin{tabular}{llccccccc}
\toprule
Training data & Encoder & AUROC & F1 (\%) & Acc. (\%) & Prec. (\%) & Rec. (\%) & Dice & mIoU \\
\midrule
0.5\% & DINOv3       & 0.605 & 47.8 & 60.8 & 70.6 & 36.1 & 0.090 & 0.047 \\
      & RETFound     & \textbf{0.766} & \textbf{72.7} & \textbf{77.1} & \textbf{89.2} & \textbf{62.3} & \textbf{0.141} & \textbf{0.076} \\
      & ImageNet ViT & 0.709 & 63.0 & 70.9 & 85.3 & 49.9 & 0.109 & 0.058 \\
\midrule
1\%   & DINOv3       & 0.588 & 45.8 & 60.5 & 72.0 & 33.5 & 0.081 & 0.042 \\
      & RETFound     & 0.740 & 67.4 & 74.2 & \textbf{90.1} & 53.8 & \textbf{0.145} & \textbf{0.078} \\
      & ImageNet ViT & \textbf{0.753} & \textbf{69.4} & \textbf{75.2} & 89.3 & \textbf{56.8} & 0.125 & 0.067 \\
\midrule
2\%   & DINOv3       & 0.597 & 46.1 & 62.2 & 78.6 & 32.6 & 0.084 & 0.044 \\
      & RETFound     & 0.759 & 71.8 & 76.4 & 88.1 & 60.6 & 0.137 & 0.074 \\
      & ImageNet ViT & \textbf{0.777} & \textbf{73.4} & \textbf{77.7} & \textbf{90.1} & \textbf{61.9} & \textbf{0.143} & \textbf{0.077} \\
\midrule
5\%   & DINOv3       & 0.655 & 55.9 & 66.1 & 79.0 & 43.2 & 0.109 & 0.058 \\
      & RETFound     & 0.762 & 72.2 & 76.9 & 89.6 & 60.4 & 0.143 & 0.077 \\
      & ImageNet ViT & \textbf{0.816} & \textbf{77.4} & \textbf{80.5} & \textbf{91.2} & \textbf{67.2} & \textbf{0.177} & \textbf{0.097} \\
\midrule
100\% & DINOv3       & 0.719 & 63.2 & 70.5 & 83.0 & 51.0 & 0.140 & 0.075 \\
      & RETFound     & 0.859 & 82.7 & 84.2 & \textbf{90.4} & 76.3 & 0.220 & 0.124 \\
      & ImageNet ViT & \textbf{0.921} & \textbf{85.9} & \textbf{87.3} & 89.1 & \textbf{84.7} & \textbf{0.242} & \textbf{0.138} \\
\bottomrule
\end{tabular}
\end{table*}
 
\subsubsection{Feature-level scoring}
To compare anomaly scoring in pixel space with scoring in feature space, we evaluated the complete models with two scoring strategies. In reconstruction-based (pixel-level) scoring, abnormality was measured from the residual between each image and its reconstruction. In feature-level scoring, the patch features of each image were compared with a bank of patch features from normal training images, and abnormality was measured as the distance to the nearest normal feature. Both strategies used the same trained models and thresholds calibrated on normal and synthetically perturbed validation images using Youden's J statistic (Table~\ref{tab:feature_level}).

\begin{table*}[t]
\centering
\small
\caption{Comparison of reconstruction-based (pixel-level) and feature-level anomaly scoring for the complete models. Image-level metrics are computed on Kermany and localization metrics on RETOUCH. mIoU is foreground-only (background excluded from the class average).}
\label{tab:feature_level}
\begin{tabular}{llccccccc}
\toprule
Scoring & Encoder & AUROC & F1 (\%) & Acc. (\%) & Prec. (\%) & Rec. (\%) & Dice & mIoU \\
\midrule
Pixel level   & DINOv3       & 0.719 & 63.2 & 70.5 & 83.0 & 51.0 & 0.140 & 0.075 \\
              & RETFound     & 0.859 & 82.7 & 84.2 & 90.4 & 76.3 & 0.220 & 0.124 \\
              & ImageNet ViT & \textbf{0.921} & \textbf{85.9} & \textbf{87.3} & 89.1 & \textbf{84.7} & 0.242 & 0.138 \\
\midrule
Feature level & DINOv3       & 0.871 & 54.0 & 68.3 & 96.7 & 37.4 & 0.273 & 0.158 \\
              & RETFound     & 0.91  & 74.4 & 79.3 & 97.8 & 60.5 & \textbf{0.397} & \textbf{0.248} \\
              & ImageNet ViT & 0.910 & 73.5 & 79.0 & \textbf{98.8} & 58.8 & 0.318 & 0.189 \\
\bottomrule
\end{tabular}
\end{table*}

\subsubsection{Perturbation strategy}

\begin{figure*}[t]
  \centering
  \includegraphics[width=\textwidth]{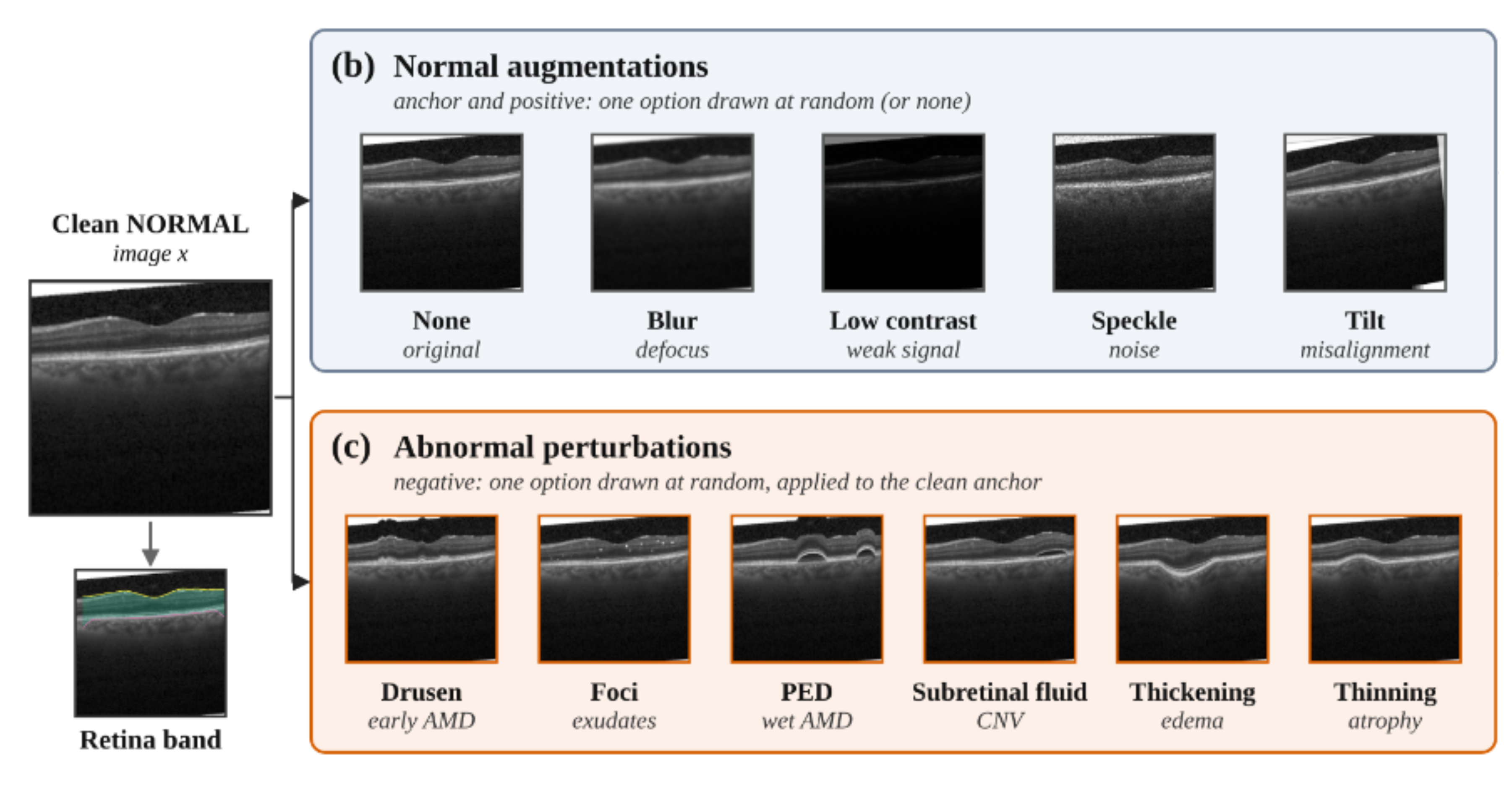}
  \caption{Perturbation strategy used to build training triplets from healthy (normal) OCT B-scans. A clean normal image $x$ is used as the anchor.
  Its retinal band is detected automatically and used to place and scale the synthetic lesions.}
  \label{fig:perturbation_strategy}
\end{figure*}

To examine the effect of different synthetic perturbations on 
anomaly detection and localization, we conducted an ablation 
study using the ImageNet-pretrained ViT-L/16 encoder. Six 
perturbation strategies were considered: retinal layer thinning, 
layer thickening, subretinal fluid, pigment epithelial detachment 
(PED), foci and as demonstrated in Fig.~\ref{fig:perturbation_strategy}. 

Each model was trained using a single 
perturbation type to generate synthetic negative samples for 
triplet learning. An additional model was trained using all six 
perturbation types to evaluate their combined effect. All models 
were evaluated for image-level anomaly detection on the Kermany 
test set and pixel-level anomaly localization on RETOUCH. 
The results are presented in Table~\ref{tab:perturbation_ablation}.

\begin{table*}[t]
\centering
\caption{Comparison of different synthetic perturbation strategies for
OCT anomaly detection and segmentation. The best result for each
metric is highlighted in bold. All values are reported as percentages. mIoU is foreground-only (background excluded from the class average).}
\label{tab:perturbation_ablation}
\resizebox{\textwidth}{!}{%
\begin{tabular}{lcccccc}
\toprule
\multirow{2}{*}{Perturbation Strategy} &
\multicolumn{4}{c}{Image-Level Anomaly Detection} &
\multicolumn{2}{c}{Pixel-Level Anomaly Segmentation} \\
\cmidrule(lr){2-5} \cmidrule(lr){6-7}
& Accuracy & F1 & AUROC & AUPRC & Dice & mIoU \\
\midrule
Thinning          & 58.00 & 33.75 & 50.91 & 60.31 & 6.80 & 3.52 \\
Thickening        & 71.27 & 63.84 & 70.01 & 77.88 & 5.97 & 3.08 \\
Subretinal fluid  & 64.59 & 52.62 & 62.68 & 70.23 & 6.85 & 3.55 \\
PED               & 65.28 & 50.60 & 63.95 & 72.26 & 6.52 & 3.37 \\
Foci              & 73.60 & 66.96 & 71.63 & 79.48 & 7.94 & 4.13 \\
drusen            & 80.49 & 76.24 & 79.57 & 86.55 & 6.84 & 3.54 \\
\midrule
All perturbations & \textbf{82.62} & \textbf{81.34} &
\textbf{85.50} & \textbf{89.28} &
\textbf{10.72} & \textbf{5.66} \\
\bottomrule
\end{tabular}%
}
\end{table*}

\section{Discussion}

In this study, we compared general-purpose and retina-specific encoders for unsupervised detection and localization of retinal abnormalities in OCT. Adapting the encoders to normal retinal anatomy with synthetic structural perturbations was the main determinant of performance and produced models that outperformed previous unsupervised methods for both detection and localization, including on images from unseen devices. The most suitable pretraining depended on the task, the scoring strategy and the amount of normal data available.

With frozen encoders, all three models missed most abnormal B-scans, and DINOv3 performed below chance. Reconstruction-based anomaly detection relies on the model failing to reconstruct structures it has not learned to represent. Large pretrained encoders, however, are designed to represent almost any visual content. A linear head trained on such features can therefore reconstruct pathological structures nearly as well as normal anatomy, and the residual no longer separates the two. A similar tendency of reconstruction models to generalize to anomalous inputs has been described for autoencoders trained on normal data\cite{gong2019memae}, and adapting pretrained features to the normal training data has been shown to improve anomaly detection over using them frozen\cite{reiss2021panda}. Benchmarks of ophthalmic classification also report consistent gains from fine-tuning over frozen features\cite{zhou2025generalist,zoellin2026benchmark}.

Adapting the encoder with the triplet objective produced the largest improvement in detection for all three encoders. We attribute this to two complementary signals. The reconstruction loss focuses the representation on the anatomy of the normal retina, while the triplet objective explicitly teaches the model that structural departures from this anatomy should be represented differently.

With full training data and reconstruction-based scoring, the conventionally pretrained ImageNet ViT performed best, followed by RETFound and DINOv3. This ranking differs from the common expectation that foundation models, and particularly domain-specific ones, provide the best starting point\cite{zhou2023retfound,moor2023generalist}. It is, however, consistent with recent benchmarks of supervised ophthalmic tasks, in which natural-image and general-purpose encoders matched or exceeded retina-specific models\cite{hou2026natural,zoellin2026benchmark}, and in which no encoder pretrained exclusively on OCT, including RETFound, ranked among the best-performing models even on OCT tasks\cite{zoellin2026benchmark}. We believe the explanation lies in how well each representation suits reconstruction-based anomaly detection rather than in its general quality. RETFound was pretrained with masked autoencoding\cite{he2022mae} on a large collection of OCT images drawn from routine clinical care, which includes eyes with retinal disease\cite{zhou2023retfound}. Its representation may therefore already encode pathological structures and help the model reconstruct them, which reduces the residual contrast between normal and abnormal images. Features of the masked-autoencoder version of RETFound have also been reported to be highly similar across different colour fundus images\cite{zhou2025generalist}, suggesting that such representations separate individual images less sharply, which may further limit how well normal and abnormal scans are distinguished before adaptation. Supervised ImageNet features, in contrast, are trained to discard image detail that is not needed for object classification. After adaptation on normal OCT, such a representation may reconstruct normal anatomy well while remaining less able to reproduce structures outside this distribution, which is exactly what reconstruction-based detection requires. DINOv3 produces rich, dense features\cite{simeoni2025dinov3}, but these features are not organized for pixel reconstruction, and a linear head may be unable to map them back to images accurately. The feature-level analysis supports this interpretation.

ROI weighting improved localization for all three encoders, while its effect on detection depended on the encoder. Without weighting, much of the reconstruction error comes from the background above and below the retina, where speckle and acquisition noise are common. Concentrating the loss on the retina makes the reconstruction more accurate within the tissue, so that the remaining residuals there are more specific to pathology. Image-level detection, however, depends on the largest residuals anywhere in the B-scan. Weighting increased recall and F1 score for the ImageNet ViT and RETFound but reduced detection performance for DINOv3. A possible explanation is that, for a representation that is already difficult to decode into pixels, shifting the reconstruction effort towards the retina reduces fidelity elsewhere enough to raise residuals in normal images as well.


The value of each pretraining regime also depended on how much normal data was available. With the smallest training set, RETFound performed best for both detection and localization, consistent with having already learned the appearance of OCT images during pretraining and with the data efficiency reported for domain-specific self-supervised pretraining in supervised settings\cite{zhou2023retfound,azizi2023remedis}. Its performance changed little until the full dataset was used, whereas the ImageNet ViT improved steadily and overtook RETFound with a small fraction of the data. RETFound may start closer to the target domain but gain less from limited adaptation, whereas the ImageNet representation must first learn OCT anatomy but then benefits more from each additional normal image. In practice, retina-specific pretraining may be most useful when a model must be adapted to a new clinical site or device with very few normal scans, while conventionally pretrained encoders may be preferable when more normal data are available. DINOv3 remained the weakest encoder at every data size, indicating that its limitation in this setting is not caused by insufficient adaptation data.

Feature-level scoring showed that the representations contain more information about abnormality than their reconstructions reveal. Measuring abnormality directly in feature space improved localization for all three encoders and improved detection AUROC for DINOv3 and RETFound. The gain was largest for DINOv3, indicating that its adapted representation does capture abnormality but that this information is largely lost when the features are mapped back to pixels by a linear head. This agrees with the strong performance of feature-based anomaly detection methods\cite{defard2021padim,roth2022patchcore} and with the design of DINOv3, which was optimized for dense feature quality rather than pixel reconstruction\cite{simeoni2025dinov3}. Feature-level scoring also changed the ranking for localization, with RETFound achieving the best localization in the study. Retina-specific pretraining therefore appears to produce patch features that distinguish normal from abnormal retinal tissue particularly well, an advantage that reconstruction-based scoring hides because RETFound can also reconstruct abnormal structures.

This study has limitations that also point to future work. First, localization was evaluated only against fluid annotations, which cover a subset of the abnormalities that the anomaly maps can highlight; datasets with annotations of drusen, atrophy and layer disruption would allow a more complete assessment. Second, we deliberately used a linear reconstruction head to make the comparison between encoders fair. This choice limits absolute performance, and stronger decoders or combinations of pixel-level and feature-level scoring may improve results. Third, we compared three encoders of a single architecture and size, and other foundation models and model sizes should be evaluated. Finally, the framework operates on individual 2D B-scans; extending it to OCT volumes and evaluating it prospectively with clinicians will be needed to establish its value for screening and triage.

In conclusion, pretrained encoders, whether general-purpose or retina-specific, did not detect retinal abnormalities without adaptation. Adapting them to normal retinal anatomy with synthetic structural perturbations produced annotation-free detectors that outperformed previous unsupervised methods for both detection and localization. These findings indicate that adaptation to normal retinal structure, rather than foundation-model  alone, is the key to building unsupervised OCT anomaly detectors.

\section{Methods}
 
 \subsection{Image preprocessing}
All B-scans were processed in the same way for both datasets. Images were resized to $256\times256$ pixels using bicubic interpolation and then normalized with the ImageNet channel mean and standard deviation. We did not apply any data augmentation during the main framework training, so that the models learned only from the natural variation present in normal B-scans.

\subsection{Pretrained encoders}
We selected three encoders that represent different pretraining strategies.
 
DINOv3\cite{simeoni2025dinov3} represents general-purpose foundation models. It was pretrained with self-supervision on a very large collection of natural images and is designed to produce dense, spatially detailed features that transfer to many tasks without labels. If broad visual knowledge is enough to capture retinal structure, this model should perform well even though it has never seen OCT images during pretraining.
 
RETFound\cite{zhou2023retfound} represents domain-specific foundation models. It was pretrained with masked autoencoding\cite{he2022mae} on a large collection of retinal OCT images and has been shown to transfer well to a range of ophthalmic tasks. It therefore tests whether prior exposure to retinal images gives an advantage over general pretraining.
 
The ImageNet-pretrained vision transformer\cite{dosovitskiy2021image} represents conventional supervised transfer learning. It was trained to classify natural images into object categories and has long been the default starting point in medical imaging. It serves as a baseline that shows how much the two foundation models add beyond standard practice.
 
Together, these models separate two factors that are often mixed in the literature. Comparing DINOv3 with RETFound, both self-supervised, shows the effect of the pretraining domain (natural images versus retinal OCT). Comparing DINOv3 with the ImageNet model, both pretrained on natural images, shows the effect of the type of pretraining (self-supervised versus supervised). To make the comparison fair, all three encoders share the same architecture, ViT-L/16. Each divides a $256\times256$ image into 256 patches of $16\times16$ pixels and represents every patch with a 1024-dimensional token. For reconstruction, we used the patch tokens from the final layer of each encoder. The class token was not used for reconstruction but provided the image-level embedding for the triplet objective. Components specific to the original pretraining task were removed: the register tokens of DINOv3, the classification head of the ImageNet model and the decoder of RETFound. For RETFound, the patch tokens were also returned to their original spatial order, because the masked autoencoder rearranges them internally. As a result, the three encoders produce features of the same shape and can be connected to the same reconstruction head.

\subsection{Reconstruction framework}
Each encoder was turned into an anomaly detector by training it to reconstruct normal B-scans. A model that has learned the appearance of the normal retina should reconstruct normal images accurately but fail to reconstruct structures it has not learned, such as fluid, drusen or neovascular lesions. The difference between an image and its reconstruction can therefore indicate whether the image is abnormal and where the abnormality is located as illustrated in Fig.~\ref{fig:framework}. To this end, we attached a reconstruction head that maps patch tokens back to pixels. Each token $z_i\in\mathbb{R}^{1024}$ was mapped independently by a single linear layer
\begin{equation}
D_{\phi}:\mathbb{R}^{1024}\rightarrow\mathbb{R}^{768},
\end{equation}
where 768 corresponds to the $16\times16\times3$ pixel values of one patch. The reconstructed patches were then placed back on a $16\times16$ grid to form the reconstructed image $\hat{x}$.
 
We kept this head as simple as possible on purpose. A deep decoder could learn to produce realistic retinal images by itself, which would hide differences between the encoders. A linear head with about 0.79 million parameters, in contrast, has no spatial context and cannot add structure that is not already present in the tokens. The quality of the reconstruction, and therefore the quality of the anomaly signal, depends mainly on the encoder.

We evaluated each encoder in four configurations of increasing adaptation. In the first, the encoder was kept frozen and only the reconstruction head was trained with the reconstruction loss, which shows how useful the original pretrained features are for OCT.  In the second, the encoder and reconstruction head were trained jointly using only the reconstruction loss, allowing us to assess the effect of reconstruction-based adaptation without triplet learning. In the third, the encoder and the head were trained together with the reconstruction loss and the triplet objective described below, which shows how much each representation benefits from adapting to normal OCT anatomy. In the fourth, a separate retinal ROI-weighted reconstruction term was added, so that reconstruction errors within the retina received greater emphasis than errors in the background. The fourth configuration is the complete model used in all further analyses.
 
\subsection{Training objectives}
The models were trained to reconstruct normal B-scans using a loss with three complementary terms:
\begin{equation}
\mathcal{L}_{\mathrm{rec}}=
\lambda_{\mathrm{pix}}\mathcal{L}_{\mathrm{pix}}+
\lambda_{\mathrm{ssim}}\mathcal{L}_{\mathrm{ssim}}+
\lambda_{\mathrm{perc}}\mathcal{L}_{\mathrm{perc}}.
\end{equation}
The pixel term $\mathcal{L}_{\mathrm{pix}}$ is an L1 loss that measures intensity differences. Because much of an OCT B-scan is background above and below the retina, the complete model gave retinal pixels a higher weight in this term. A binary retinal mask $M$ was obtained from the clean target image using multi-scale horizontal Gabor filters, which respond strongly to the layered structure of the retina, followed by binarization, as in our previous study\cite{haghighi2026anatomy}. The detailed ROI extraction
procedure and spatial weighting configuration are provided
in Supplementary Section~S2. Each pixel $i$ was weighted by $1+\alpha M_i$, and the weights were normalized to a mean of one in each image; $\alpha=0$ corresponds to the configuration without ROI weighting. The structural term $\mathcal{L}_{\mathrm{ssim}}=1-\mathrm{SSIM}(x,\hat{x})$ encourages the reconstruction to preserve local contrast and layer organization. The perceptual term $\mathcal{L}_{\mathrm{perc}}$ compares the original and reconstructed images at the level of features rather than pixels. It is the L1 distance between the patch tokens of $x$ and $\hat{x}$, computed by a frozen copy of the same pretrained encoder used in the model. Using each model's own pretrained features, rather than a single shared network, prevents the perceptual loss from favoring one encoder over the others.
 
A model trained only to reconstruct normal images learns what normal anatomy looks like, but it is never told which changes in structure should be treated as abnormal. It may therefore also learn to reconstruct abnormal structures well, which would weaken the anomaly signal. To address this, we added the triplet objective, which provides an explicit signal about structural abnormality without using any real pathology. Each training triplet contained an anchor $x_a$, which was a normal B-scan; a positive $x_p$, which was a different normal B-scan; and a negative $x_n$, which was a synthetically perturbed version of the anchor. The anchor and positive were reconstructed to themselves, and the negative was reconstructed towards the clean anchor. The triplet loss pulls the representations of normal images together and pushes the perturbed image away from its normal source:
\begin{equation}
\mathcal{L}_{\mathrm{tri}}=
\max\left(
\lVert f(x_a)-f(x_p)\rVert_2^2-
\lVert f(x_a)-f(x_n)\rVert_2^2+m,\,0
\right),
\end{equation}
where $m=0.1$ is the margin and $f(\cdot)$ is an image-level embedding obtained by projecting the encoder's class token to 128 dimensions with a linear layer, followed by L2 normalization. The full training objective is
\begin{equation}
\mathcal{L}=\mathcal{L}_{\mathrm{rec}}+\lambda_{\mathrm{tri}}\mathcal{L}_{\mathrm{tri}}.
\end{equation}
We used the triplet objective as a fixed adaptation step, applied in the same way to every encoder, so that we could compare how different representations respond to the same signal.
 
The synthetic negatives were created from normal B-scans using the two perturbation families. Structural deformation locally thickens and bends the retinal layers, and localized intensity perturbation changes the brightness of a small retinal region. Both were applied only within the retina, and each anchor was paired with a single negative generated by randomly selecting one of the two perturbation families. These perturbations were not designed to imitate any particular disease. Instead, they represent general types of structural change, so that the model learns to separate normal anatomical variation from departures from it. The complete training procedure is summarized in Algorithm~\ref{alg:training}.
 
\begin{algorithm}[t]
\caption{Structure-aware adaptation of a pretrained encoder}
\label{alg:training}
\begin{algorithmic}[1]
\Require normal OCT training set $\mathcal{D}$; pretrained ViT-L/16 encoder $E$; linear reconstruction head $D_{\phi}$; linear projection head $g$; frozen copy of the pretrained encoder $F_E$; ROI weight $\alpha$ ($\alpha=0$ without ROI weighting)
\For{each training iteration with mini-batch $\mathcal{B}\subset\mathcal{D}$}
  \For{each anchor $x_a\in\mathcal{B}$}
    \State Sample a different normal B-scan $x_p\sim\mathcal{D}$, $x_p\neq x_a$
    \State $M_a\gets\Call{GaborROI}{x_a}$ \Comment{retinal mask of the anchor}
    \State $x_n^{\mathrm{def}}\gets\Call{Deform}{x_a,M_a}$;\quad $x_n^{\mathrm{int}}\gets\Call{IntensityPerturb}{x_a,M_a}$
    \For{$(x,x_t)\in\{(x_a,x_a),(x_p,x_p),(x_n^{\mathrm{def}},x_a),(x_n^{\mathrm{int}},x_a)\}$}
      \State $Z\gets E(x)$ \Comment{256 patch tokens}
      \State $\hat{x}\gets\Call{Unpatchify}{D_{\phi}(Z)}$
      \State $M_t\gets\Call{GaborROI}{x_t}$ \Comment{mask from the clean target}
      \State Compute $\mathcal{L}_{\mathrm{rec}}(x_t,\hat{x};M_t,\alpha)$ \Comment{L1 + SSIM + perceptual}
    \EndFor
    \State $f(x)\gets g(E_{\mathrm{cls}}(x))\,/\,\lVert g(E_{\mathrm{cls}}(x))\rVert_2$ for $x\in\{x_a,x_p,x_n^{\mathrm{def}},x_n^{\mathrm{int}}\}$
    \State Compute $\mathcal{L}_{\mathrm{tri}}(f(x_a),f(x_p),f(x_n))$ for $x_n\in\{x_n^{\mathrm{def}},x_n^{\mathrm{int}}\}$
  \EndFor
  \State Update $E$, $D_{\phi}$ and $g$ using $\mathcal{L}=\mathcal{L}_{\mathrm{rec}}+\lambda_{\mathrm{tri}}\mathcal{L}_{\mathrm{tri}}$ \Comment{$F_E$ stays frozen}
\EndFor
\end{algorithmic}
\end{algorithm}
 
\subsection{Image-level anomaly scoring}
After training, each test image was passed through the model and compared with its reconstruction using the absolute residual map $R=|x-\hat{x}|$. Many retinal lesions occupy only a small part of a B-scan, and averaging the residual over the whole image would dilute their signal. We therefore defined the anomaly score $A(x)$ as the mean of the 5\% largest residuals in the image.
 

To classify images without using any real pathological data, we set the decision threshold T using normal and synthetically perturbed validation images. The threshold was selected by maximizing Youden's $J$ statistic, defined as $J(t)=\mathrm{Sensitivity}(t)+\mathrm{Specificity}(t)-1$, where normal images were treated as negative samples and synthetically perturbed images as positive samples. The selected threshold was then applied to the independent test set without further adjustment. A test image was classified as abnormal when $A(x) > T$. For each model, we kept the checkpoint with the lowest loss on the normal validation set.
 
\subsection{Anomaly localization}
Beyond deciding whether an image is abnormal, a clinically useful detector should indicate where the abnormality is. Here, localization means identifying the pathological region responsible for the anomaly signal, not segmenting the retinal layers. We used the residual map $R$ as a pixel-level anomaly map and applied the models trained on Kermany directly to RETOUCH, without further training or recalibration. To obtain a binary prediction, we used the 99th percentile of pixel residuals in the Kermany normal validation images as the threshold, so that no RETOUCH information was involved. The predicted regions were compared with the combined IRF, SRF and PED annotations using the Dice coefficient and mIoU.
 
Because all models were trained on images from a single device, whereas RETOUCH includes three devices that differ in resolution, noise and intensity, this evaluation also tests whether the anomaly maps localize pathology on devices that were not seen during training.
 
\subsection{Feature-level anomaly scoring}
The reconstruction-based score measures abnormality in pixel space, after the encoder features have passed through the linear head. Because this head is deliberately simple, some information that the encoder has captured about abnormal structure may not appear in the reconstruction. To measure abnormality directly in the representation, we also scored images in feature space.
 
Following memory-bank approaches to anomaly detection\cite{roth2022patchcore}, we extracted the patch features of the normal training images for each trained model and stored them in a feature bank. For a test image, each patch feature was compared with the bank, and its anomaly score was the distance to its nearest normal feature. A large distance means that the patch does not resemble any normal patch seen during training. The image-level score was obtained by aggregating the highest patch scores, and the decision threshold was calibrated on normal and synthetically perturbed validation images using Youden's J statistic, in the same way as for reconstruction-based scoring. For localization, the patch scores were kept separately, arranged on the $16\times16$ patch grid and resized to the image resolution, so that patches far from all normal features formed the predicted anomaly region. As with the reconstruction-based approach, no real pathological image or annotation was used to build the feature bank or set the thresholds. Additional details of normal feature-bank construction
and nearest-neighbor anomaly scoring are provided in
Supplementary Section~S4.

\subsection{Ablation studies}

Four ablation studies were performed to examine the influence of training objectives, training data availability, anomaly scoring strategies and synthetic perturbations on detection and localization performance.


The first ablation investigated the contribution of different training components to the learned representations. Four configurations were considered. In the first, the pretrained encoder was frozen and only the linear reconstruction head was trained, allowing the reconstruction capability of the original pretrained features to be assessed without modifying the encoder. In the second, the encoder and reconstruction head were jointly optimized using only the reconstruction loss to assess reconstruction-based adaptation without triplet learning. In the third, the encoder and reconstruction head were jointly optimized using reconstruction and triplet losses to examine the effect of adapting the representation to normal retinal anatomy while distinguishing synthetic abnormalities. The fourth configuration introduced ROI weighting to the reconstruction loss to assess whether emphasizing retinal structures, rather than treating retinal and background pixels equally, improves anomaly detection and localization.

The second ablation examined the sensitivity of the framework to the
amount of available normal training data. Because the proposed approach relies exclusively on normal images, understanding its data requirements is important for settings where large collections of healthy OCT scans may not be available. The complete models were trained using randomly selected subsets containing 0.5\%, 1\%, 2\% and 5\% of the normal training set and compared with models trained on the full dataset.
These fractions were selected to examine performance under highly
restricted training conditions and how it changes as more normal
samples become available. All other training settings were kept fixed, and each subset size was repeated with ten independent random subsets to account for variability in sample selection.

The third ablation compared reconstruction-based and feature-level
anomaly scoring to examine how the representation of abnormalities
affects detection and localization. Reconstruction-based scoring
identifies abnormalities through pixel-level differences between
the input and its reconstruction, whereas feature-level scoring
measures deviations from the representations of normal retinal
images using a normal feature bank. This comparison was motivated
by the possibility that structural changes captured by the encoder
may not be fully reflected in the reconstructed image, particularly
when using a simple linear decoder. Both strategies were applied
to the same trained models, allowing the effect of the scoring
approach to be examined independently of model training.

The fourth ablation examined how the type and diversity of synthetic
perturbations used to construct training triplets affect anomaly detection and localization. Six pathology-inspired perturbations were selected to represent different changes in retinal anatomy: thinning and thickening capture changes in retinal layer geometry, drusen and PED introduce localized elevations and deformations, subretinal fluid creates hyporeflective spaces,
and foci introduce localized hyperreflective regions. These perturbations were chosen to examine how different structural and intensity changes contribute to representation learning without requiring pathological training images. Two settings were compared: a combined strategy using all six perturbations with normal augmentations (blur, low contrast, speckle or tilt) applied
to anchors and positives; and six single-perturbation models, each trained using only one abnormal perturbation while retaining the same normal augmentations (Fig.~\ref{fig:perturbation_strategy}). The perturbation generation settings, including the ranges
used to vary lesion size, number and location, are provided
in Supplementary Section~S3. All models were selected using held-out normal validation data and evaluated on the same Kermany and RETOUCH datasets.

\subsection{Comparison methods}
We compared the pretrained encoders with six unsupervised methods that were trained and evaluated using identical data partitions and evaluation protocol: a ConvNeXt-based reconstruction model\cite{liu2022convnext}, a variational autoencoder (VAE)\cite{kingma2014vae}, VQ-VAE\cite{vandenoord2017vqvae}, VQGAN\cite{esser2021vqgan}, f-AnoGAN\cite{schlegl2019fano} and an anatomy-aware VQGAN that combines a vector-quantized generative model with the triplet objective and ROI weighting\cite{haghighi2026anatomy}.
 
\subsection{Evaluation metrics}
The primary metric for image-level detection was AUROC, which summarizes performance across all thresholds. At the threshold $T$, we also report accuracy, F1 score, precision and recall. For localization, we report the Dice coefficient and mIoU.
 
\subsection{Implementation}
All models were implemented in PyTorch and trained on a single GPU with 48\,GB of memory. Detailed model configurations,
optimization hyperparameters and loss weights are provided
in Supplementary Section~S1.

 \subsection{Statistical analysis}

Image-level anomaly detection was evaluated on the held-out Kermany test set using AUROC, accuracy, and F1 score. Pixel-level localization was evaluated on RETOUCH using the Dice coefficient and mIoU. All models were evaluated on identical test sets to ensure consistent comparisons. For image-level detection, decision thresholds were determined using Youden's J statistic on normal and synthetically perturbed Kermany validation images. The threshold maximizing the difference between sensitivity and false-positive rate was selected for each model and scoring strategy and applied unchanged to the independent test set. For reconstruction-based pixel-level localization, the threshold was set at the 99th percentile of pixel-level reconstruction errors from normal Kermany validation images and applied unchanged to RETOUCH. No real pathological images or annotations were used for threshold calibration.

Statistical uncertainty was quantified using 100 bootstrap resamples,
with 95\% confidence intervals calculated from the 2.5th
and 97.5th percentiles of the bootstrap distributions.
Differences between models were assessed using paired
bootstrap comparisons, in which the same resampled test
observations were used to calculate the performance
difference between models in each iteration. For the
data-efficiency ablation, each training fraction was evaluated
using ten independently sampled subsets, with performance
summarized as the mean.

\printbibliography
\section*{Acknowledgements}
This work was supported by the U.S. National Institutes of Health and National Eye Institute under Grants 1R01EY037828-01, R15EY035804 and R21EY035271.

\section*{Author Information}
\subsection*{contributions}
T.H. and S.G. conceived the study and developed the experimental framework. T.H. and S.G. implemented the models, conducted experiments, and analyzed results. H.T. contributed to study design and methodological interpretation. M.N.A. supervised the project, contributed to the clinical and technical framing, and acquired funding. T.H. and S.G. drafted the manuscript. All authors reviewed and revised the manuscript and approved the submitted version.

\subsection*{Corresponding author}
Correspondence to \href{mailto:malam8@charlotte.edu}{Minhaj Nur Alam}.

\section*{Competing interests}
The authors declare no competing interests.

\section*{Data availability}
The Kermany OCT dataset is publicly available from Mendeley Data (\url{https://data.mendeley.com/datasets/rscbjbr9sj}). The RETOUCH dataset is available from the challenge website (\url{https://retouch.grand-challenge.org}) under its data-use terms.

\section*{Code availability}
The code for preprocessing, training, inference and evaluation will be released in a public repository upon publication.

\clearpage

\clearpage
\resetlinenumber[1]
\linenumbers

\setcounter{figure}{0}
\setcounter{table}{0}
\setcounter{equation}{0}
\renewcommand{\thefigure}{S\arabic{figure}}
\renewcommand{\thetable}{S\arabic{table}}
\renewcommand{\theequation}{S\arabic{equation}}
\makeatletter
\@ifpackageloaded{hyperref}{%
  \renewcommand{\theHfigure}{supp.\arabic{figure}}%
  \renewcommand{\theHtable}{supp.\arabic{table}}%
  \renewcommand{\theHequation}{supp.\arabic{equation}}%
}{}
\makeatother

\section*{Supplementary Information}
\addcontentsline{toc}{section}{Supplementary Information}

\subsection*{S1. Model configuration and training}
All models used a ViT-L/16 encoder with a lightweight linear patch decoder.
The final-layer patch representations were mapped independently from 1,024
features to $16\times16\times3$ reconstructed pixel values. A separate linear
projection mapped the class token to a 128-dimensional, $\ell_2$-normalized
embedding for the triplet objective. The same architecture and training
protocol were applied to DINOv3, RETFound and the supervised ImageNet ViT.
The principal settings are summarized in Table~\ref{tab:supp_train_config}.

\begin{table}[htbp]
\centering
\small
\caption{Model architecture, training hyperparameters,
and optimization settings used for the OCT anomaly
detection framework.}
\label{tab:supp_train_config}
\begin{tabular}{@{}p{0.46\linewidth}p{0.45\linewidth}@{}}
\toprule
Parameter & Setting \\
\midrule
\multicolumn{2}{@{}l}{\textit{Model architecture and preprocessing}} \\
Pretrained encoders & DINOv3, RETFound, and supervised ImageNet ViT-L/16 \\
Input resolution & $256\times256$ pixels \\
Image preprocessing & Bicubic resizing; ImageNet normalization \\
Patch size and grid & $16\times16$ pixels; $16\times16$ patches (256 tokens) \\
Patch embedding dimension & 1,024 \\
Reconstruction decoder & Single linear layer, $1024\rightarrow768$ \\
Triplet projection head & Linear, $1024\rightarrow128$; $\ell_2$ normalization \\

\addlinespace
\multicolumn{2}{@{}l}{\textit{Optimization}} \\
Optimizer & AdamW \\
Batch size & 16 \\
Training epochs & 100 \\
Encoder learning rate & $1\times10^{-5}$ \\
Decoder learning rate & $1\times10^{-3}$ \\
Projection learning rate & $1\times10^{-4}$ \\
Weight decay & $1\times10^{-4}$ \\
Precision & Mixed precision (FP16) \\

\addlinespace
\multicolumn{2}{@{}l}{\textit{Training objectives and selection}} \\
Reconstruction loss weights (L1, SSIM, perceptual) & $1.0$, $0.2$, $0.1$ \\
Triplet loss weight & $0.5$ \\
Triplet margin & $0.1$ \\
ROI weighting factor & $\alpha=4.0$; spatial weights normalized to unit mean \\
Model selection & Lowest loss on normal-only validation data \\
Training hardware & Single GPU with 48\,GB memory \\
\bottomrule
\end{tabular}
\end{table}

\subsection*{S2. Retinal region-of-interest estimation}
The retinal ROI was estimated automatically from each clean OCT B-scan.
Bright image borders were
removed before intensity normalization and gamma correction ($\gamma=0.7$).
Horizontal retinal structures were enhanced using a bank of Gabor filters
with a $31\times31$ kernel, wavelengths $\{4,8,16\}$ pixels, scales
$\sigma\in\{2,4\}$ and orientation $0^\circ$. Maximum filter responses were
combined and binarized with Otsu thresholding, followed by morphological
opening and closing. A connected component spanning a substantial portion of
the image width was selected to identify the retinal band.

The resulting mask $M$ was used to emphasize the retinal tissue relative to
background in the reconstruction objective. With $M_i\in\{0,1\}$ and ROI
weighting factor $\alpha$, the spatial weights were defined by
\begin{equation}
 w_i = \frac{1+\alpha M_i}{\frac{1}{N}\sum_{j=1}^{N}(1+\alpha M_j)}.
\end{equation}

\subsection*{S3. Synthetic perturbation generation}

Synthetic negative samples were constructed from normal OCT B-scans to supply
a controlled abnormality signal without requiring diseased training images.
The six perturbation families evaluated in the perturbation ablation were
selected to span complementary anatomical changes. Perturbation placement
was guided by the ROI of the clean image so that modifications affected
the retinal structures rather than unrelated background pixels.

The size, number, and location of synthetic abnormalities were varied within predefined ranges based on image width and retinal thickness. Local retinal thickness was reduced by 20–40\% for thinning and increased by 25–60\% for thickening. Drusen and hyperreflective foci were generated as 3–9 and 8–25 localized abnormalities per image, respectively. Subretinal fluid and PED were simulated by introducing localized retinal elevations with varying widths and heights.

\subsection*{S4. Normal feature bank and feature-level scoring}
To evaluate anomaly evidence directly in feature space, a normal reference
bank was constructed independently for each trained encoder. Final-layer
spatial patch embeddings were extracted from normal Kermany training images
and stored as a collection $\mathcal{B}$ of reference features. For each test
patch embedding $z_i$, the patch anomaly score was defined as its distance to
the nearest normal reference feature:
\begin{equation}
 a_i=\min_{b\in\mathcal{B}}d(z_i,b).
\end{equation}
Here, $d$ denotes the distance used by the feature-bank implementation.
Higher values indicate stronger deviations from the representations observed
in normal training images. Patch scores were arranged according to their
spatial locations and upsampled to the image resolution to obtain a dense
anomaly map. Image-level scores were obtained by aggregating the most anomalous
patches.

\end{document}